\documentclass[conference]{IEEEtran}

\IEEEoverridecommandlockouts
\usepackage{cite}
\usepackage[numbers,sort&compress]{natbib}

\usepackage[utf8]{inputenc}
\usepackage{amsmath,amssymb}
\usepackage{graphicx}
\usepackage{bm}
\usepackage{enumerate}
\usepackage{comment}
\usepackage{xcolor}
\usepackage{url}
\usepackage{color,soul}
\usepackage{subcaption}
\usepackage{float}
\usepackage{tablefootnote}

\usepackage{subfig}
\usepackage{booktabs}

\let\labelindent\relax
\usepackage[shortlabels]{enumitem}

\definecolor{light-gray}{gray}{0.8}

\usepackage{amsmath,amssymb,amsfonts}
\usepackage{algorithmic}
\usepackage{graphicx}
\usepackage{textcomp}
\usepackage{xcolor}
\usepackage{subcaption}
\usepackage{lipsum}
\usepackage{multirow}

\begin{document}

\title{Enhance Image-to-Image Generation with LLaVA-generated Prompts \\
}

\author{
\small 

\begin{tabular}[t]{c@{\extracolsep{8em}}c} 

1\textsuperscript{st} Zhicheng Ding\footnotemark & 2\textsuperscript{nd} Panfeng Li \\
\textit{Fu Foundation School of Engineering and Applied Science} & \textit{Department of Electrical and Computer Engineering} \\
\textit{Columbia University} & \textit{University of Michigan} \\
New York, USA & Ann Arbor, USA \\
zhicheng.ding@columbia.edu & pfli@umich.edu \\

\\

3\textsuperscript{rd} Qikai Yang & 4\textsuperscript{th} Siyang Li \\
\textit{Department of Computer Science} & \textit{Lubin School of Business} \\
\textit{University of Illinois Urbana-Champaign} & \textit{Pace University} \\
Urbana, USA & New York, USA \\
qikaiy2@illinois.edu & lisiyang98@hotmail.com \\

\end{tabular}

}

\maketitle

\begin{abstract}
    This paper presents a novel approach to enhance image-to-image generation by leveraging the multimodal capabilities of the Large Language and Vision Assistant (LLaVA). We propose a framework where LLaVA analyzes input images and generates textual descriptions, hereinafter LLaVA-generated prompts. These prompts, along with the original image, are fed into the image-to-image generation pipeline. This enriched representation guides the generation process towards outputs that exhibit a stronger resemblance to the input image. Extensive experiments demonstrate the effectiveness of LLaVA-generated prompts in promoting image similarity. We observe a significant improvement in the visual coherence between the generated and input images compared to traditional methods. Future work will explore fine-tuning LLaVA prompts for increased control over the creative process. By providing more specific details within the prompts, we aim to achieve a delicate balance between faithfulness to the original image and artistic expression in the generated outputs.
\end{abstract}

\begin{IEEEkeywords}
Large Language and Vision Assistant; Image-to-Image Generation; Large Language Model
\end{IEEEkeywords}

\section{INTRODUCTION}
\label{sec:intro}  

The field of artificial intelligence is experiencing a surge in the development of multifaceted assistants capable of solving many real-life problems. Large language models play an important part of it, showing their great in multiple fields such as natural language processing~\cite{pan2024chain, zhang2022can, zeng2024wordepth} and computer vision~\cite{liu2024adaptive100, shen2024localization, liu2024using, ding-19-intelligent, zhu2024cross, li2024utilizing, ge2023encouraging, zhu2021pseudo, ge2020zero}. Among those fields, one such example is Stable Diffusion~\cite{rombach2021highresolution}, a powerful model that generates novel and imaginative images based on user-provided starting points. However, relying solely on the input image presents limitations. The generated image may deviate significantly from the user's intent, resulting in outputs that lack the desired level of control or fidelity. In addition, the inaccuracy and instability of generated images from large language models are also problems observed in past studies that necessitate careful consideration and mitigation strategies moving forward~\cite{liu2023llava}.

To address the problems mentioned as above, this paper explores a novel approach that bridges this gap and empowers Stable Diffusion for more precise image generation by incorporating the capabilities of the Large Language and Vision Assistant (LLaVA). LLaVA excels in image understanding, allowing it to analyze existing images and extract key features and concepts. This visual interpretation capability can be leveraged to generate accurate and detailed textual prompts that inform the image generation process.

We propose a novel framework where LLaVA analyzes an input image and generates a refined text prompt, along with a negative prompt, that captures the essence of the original image. This enhanced prompt is then fed into Stable Diffusion, guiding it to create a new image that faithfully reflects the content and style of the original image, potentially with creative variations. By incorporating LLaVA's image understanding capabilities, this approach aims to address the shortcomings of solely image-based generation, leading to outputs that exhibit a higher degree of similarity to the user's initial input.

This paper contributes to the field of AI-powered image-to-image generation~\cite{minaee2024large, saharia2022photorealistic}. We seek to advance the state-of-the-art in AI-powered image generation, enabling users to achieve a higher degree of control and fidelity in the creative process in several keyways.

\begin{itemize}[itemsep=0em,topsep=0pt,parsep=0pt,partopsep=0pt,leftmargin=*,labelindent=5pt]
  \item \textbf{Integration Exploration}: We explore the integration of LLaVA's image understanding capabilities with Stable Diffusion's image generation prowess.
  \item \textbf{Refined Prompting Framework}: We introduce a framework for refining text and negative prompts based on the visual content of the input image.
  \item \textbf{Experimental Evaluation}: We conduct extensive experiments to evaluate the impact of LLaVA-generated prompts on the quality and similarity of image-to-image generation.
\end{itemize}

\section{METHODS}
\label{sec:method}

\begin{figure*}[ht]
	\centering
    \includegraphics[width=0.55\linewidth]{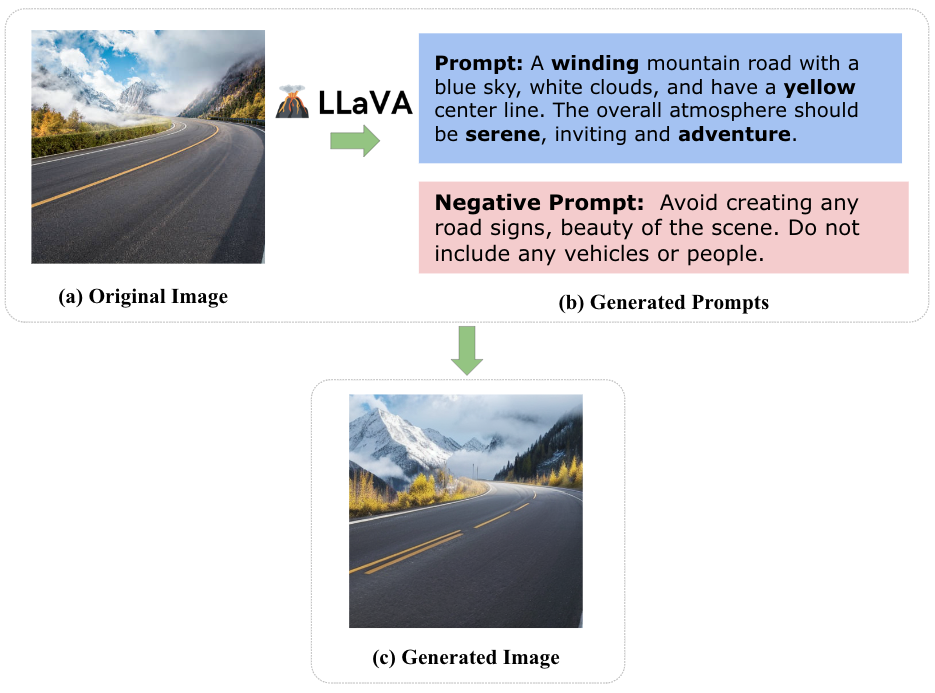}
    \caption{Framework of LLaVA-prompts-based image-to-image generation}
    \label{fig:results}
\end{figure*}

Figure [1] showcases the framework of our proposed approach, which utilizes LLaVA-generated prompts for image-to-image generation. This method deviates from directly feeding Stable Diffusion with input images. Instead, it seamlessly integrates prompts to enhance the generation process~\cite{pan2023ising, ma2022learning, wang2024research}. The workflow starts with feeding the input image (Figure 1(a)) to LLaVA. We query LLaVA to "generate prompt and negative prompt for this image." Subsequently, LLaVA generates a prompt and a negative prompt (Figure 1(b)) tailored to the specific content of the input image. Finally, both the generated prompts and the original image are fed into the image-to-image generation model (e.g., Stable Diffusion). The model then takes all inputs into consideration when generating the final output image (Figure 1(c)). The following sections will dive into the greater details of how each module works.

\subsection{Prompt Generation using LLaVA}

\begin{table*}[t]
  \centering
  \renewcommand{\arraystretch}{1.5} 
  \setlength{\tabcolsep}{8pt} 
  \begin{tabular}{*{7}{c}}
    \hline
    Images & RMSE ↓ & PSNR ↑ & FSIM ↑ & SSIM ↑ & UIQ ↑ & SRE ↑ \\
    \hline
    w/o prompt & 0.01931 & 34.2736 & 0.28770 & 0.78507 & 0.03133 & 48.7848 \\
    \hline
    with prompts & \bf{0.01008} & \bf{39.8750} & \bf{0.36375} & \bf{0.92199} & \bf{0.07616} & \bf{51.6555} \\
    \hline
  \end{tabular}
  \caption{Metric comparison between image-to-image generation with and without LLaVA generated prompts}
\end{table*}

\begin{table*}[t]
  \centering
  \renewcommand{\arraystretch}{1.5} 
  \setlength{\tabcolsep}{8pt} 
  \begin{tabular}{*{8}{c}}
    \hline
    Images & Approach & RMSE ↓ & PSNR ↑ & FSIM ↑ & SSIM ↑ & UIQ ↑ & SRE ↑ \\
    \hline
    \multirow{2}{*}{dog} & w/o prompt & 0.01952 & 34.1125 & 0.28610 & 0.78434 & 0.00179 & 52.8297 \\
    & w prompt (ours) & \bf{0.01265} & \bf{37.9437} & \bf{0.39774} & \bf{0.88115} & \bf{0.05121} & \bf{55.0153} \\
    \hline
    \multirow{2}{*}{astronaut} & w/o prompt & 0.01798 & 34.9009 & 0.27823 & 0.79330 & 0.00104 & 51.0724 \\
    & w prompt (ours) & \bf{0.01262} & \bf{37.9785} & \bf{0.30214} & \bf{0.88230} & \bf{0.02420} & \bf{52.6086} \\
    \hline
    \multirow{2}{*}{plane} & w/o prompt & 0.01275 & 37.7418 & 0.31769 & 0.92368 & 0.00861 & 53.8197 \\
    & w prompt (ours) & \bf{0.00767} & \bf{42.2708} & \bf{0.40353} & \bf{0.96597} & \bf{0.10995} & \bf{55.7809} \\
    \hline
    \multirow{2}{*}{skyscraper} & w/o prompt & 0.02603 & 31.6809 & 0.27245 & 0.58362 & -0.00236 & 43.1745 \\
    & w prompt (ours) & \bf{0.01839} & \bf{34.6985} & \bf{0.31015} & \bf{0.74276} & \bf{0.03159} & \bf{44.6849} \\
    \hline
  \end{tabular}
  \caption{Extensive experiment of comparison with and without LLaVA generated prompts}
\end{table*}

This study investigates the potential of LLaVA, a powerful large language model (LLM) known for its proficiency in understanding and describing visual content, for multimodal prompt generation~\cite{hu2024instructimagen, ramesh2021zeroshot}. We explore a novel pipeline that use LLaVA as an image interpreter that generates the prompts to enhance image-to-image generations. We will also consider combining SVM~\cite{liu2024enhanced}, a popular and efficient machine learning method, to boost the performance of our model. The core concept involves presenting LLaVA with an original image along with a task instruction. In our experiment, the instruction was "Generate prompt and negative prompt for this image." This text serves a dual purpose: (1) it defines the specific task for LLaVA and (2) guides its analysis of the image by explicitly stating the desired output format. While the descriptions generated by LLaVA can be lengthy, there is a possibility that they might not fully capture all the necessary details required for an effective textual prompt. Our findings suggest that incorporating additional, more specific task instructions within the input text could significantly enhance the precision and focus of the generated prompts.

\subsection{Image-to-Image Generation with Prompts}
Our methodology utilizes the StableDiffusionImg2ImgPipeline, a component within Stable Diffusion 2.0. This pipeline employs a diffusion-denoising mechanism, granting us fine-grained control over the image editing process. We incorporate input images alongside prompts generated by LLaVA. These prompts aim to capture the essence of the input image. Additionally, negative prompts are employed to guide the model away from unintended visual elements. By combining the input image, LLaVA-generated prompts, and negative prompts, we instruct the image-to-image pipeline to generate a new image that closely resembles the original input image but incorporates the specified modifications~\cite{he2023network, jiashu2021performance}. The fine-grained control offered by text prompts and the integration with LLaVA-generated prompts provide a robust approach for manipulating and generating new visuals~\cite{zou2021interpreting, hang2023efficient, 10402659, wang2023diffusiondb}.

\section{EXPERIMENTAL SETUP}
\label{sec:exp}

\begin{figure*}[ht]
	\centering
    \includegraphics[width=0.8\linewidth]{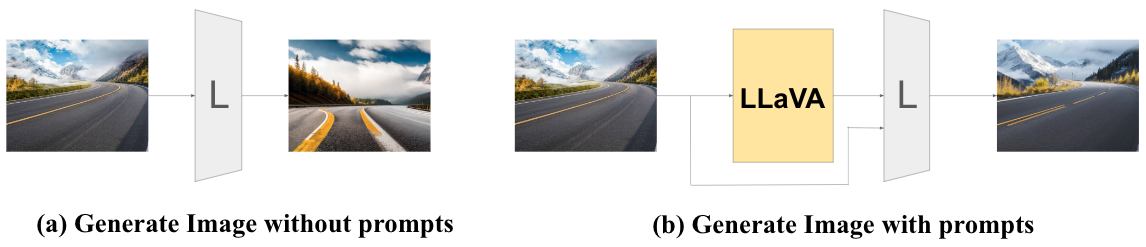}
    \caption{Comparison between image-to-image generation with and without LLaVA generated prompts}
    \label{fig:results}
\end{figure*}

\subsection{Prompt Generation using LLaVA}
In this research, we utilized a pre-trained Large Language Model (LLM), specifically the LLaVA v1.6-34b model, to generate both positive and negative prompts based on input images. Figure 1(b) showcases an example of LLaVA's output for a specific input image. The analysis of the generated prompts provides insights into LLaVA's capability to understand visual content and translate it into textual descriptions. In this case, the positive prompt captures key aspects of the image: 

\begin{itemize}[itemsep=0em,topsep=0pt,parsep=0pt,partopsep=0pt,leftmargin=*,labelindent=5pt]
  \item \textbf{Overall Ambiance}: The scene should evoke a sense of serenity.
  \item \textbf{Background Details}: The bridge is surrounded by lush and towering mountains.
  \item \textbf{Excluded Elements}: People, boats, and other similar objects should be absent from the generated image.
\end{itemize}

However, it is important to acknowledge potential limitations in the generated negative prompts. The example in Figure 1(b) might contain misleading elements. The SVM~\cite{liu2024enhanced} approach from this article helped design our experimental methods and the model can solve the problems like constraints. Further investigation is necessary to understand the effectiveness and accuracy of LLaVA's negative prompt generation in various scenarios (detailed in Future Work). Overall, LLaVA shows great potential as a tool for generating text prompts for image-to-image generation tasks.

\subsection{Image-to-Image Generation with Prompts}
This study utilizes the pre-trained open-source Stable Diffusion v2 model for image generation based on user-provided input images and prompts. Figure 1(c) presents a sample output of the generated image. The generated image exhibits a high degree of correspondence with the prompts and maintains a strong visual similarity to the input image. The subsequent section will conduct a comparative analysis of generated images produced with and without prompts generated by the LLaVA model. This comparison aims to further elucidate the advantages of our proposed approach.

\subsection{Image-to-Image Generation without Prompts}
Similar to previous section, we use the open source Stable Diffusion v2 model~\cite{rombach2021highresolution} to generate the image solely on an input image, without the incorporation of additional prompts. Figure 2(b) presents a representative example of the pipeline's output. While the generated image exhibits a high degree of stylistic and thematic similarity to the input image, it introduces a degree of noise. Specifically, the generated image deviates from the input in the following aspects: (1) the presence of two yellow lines on the road (as opposed to a single line in the original), (2) the transformation of the single-lane road into a dual-lane road, and (3) the diminished visibility of the background mountains. These observations warrant further investigation into the noise generation characteristics of the pipeline.

\subsection{Comparison w/wo Prompt}

Figure 2 (b) and (c) visually demonstrate the distinction between images generated with and without LLaVA prompts. To quantify this observed difference in image similarity, we calculate the image similarity metrics, including Root Mean Square Error (RMSE), Peak Signal-to-Noise Ratio (PSNR), Feature-based Similarity Index (FSIM), Structural Similarity Index (SSIM), Universal Image Quality Index (UIQ), and Signal to Reconstruction Error Ratio (SRE)~\cite{pan2024cormf}. The results, presented in Table 1, substantiate the qualitative observations, indicating that LLaVA-generated prompts contribute to the generation of more similar images.

\subsection{Extensive Experiments}

\begin{figure*}[ht]
	\centering
    \includegraphics[width=1\linewidth]{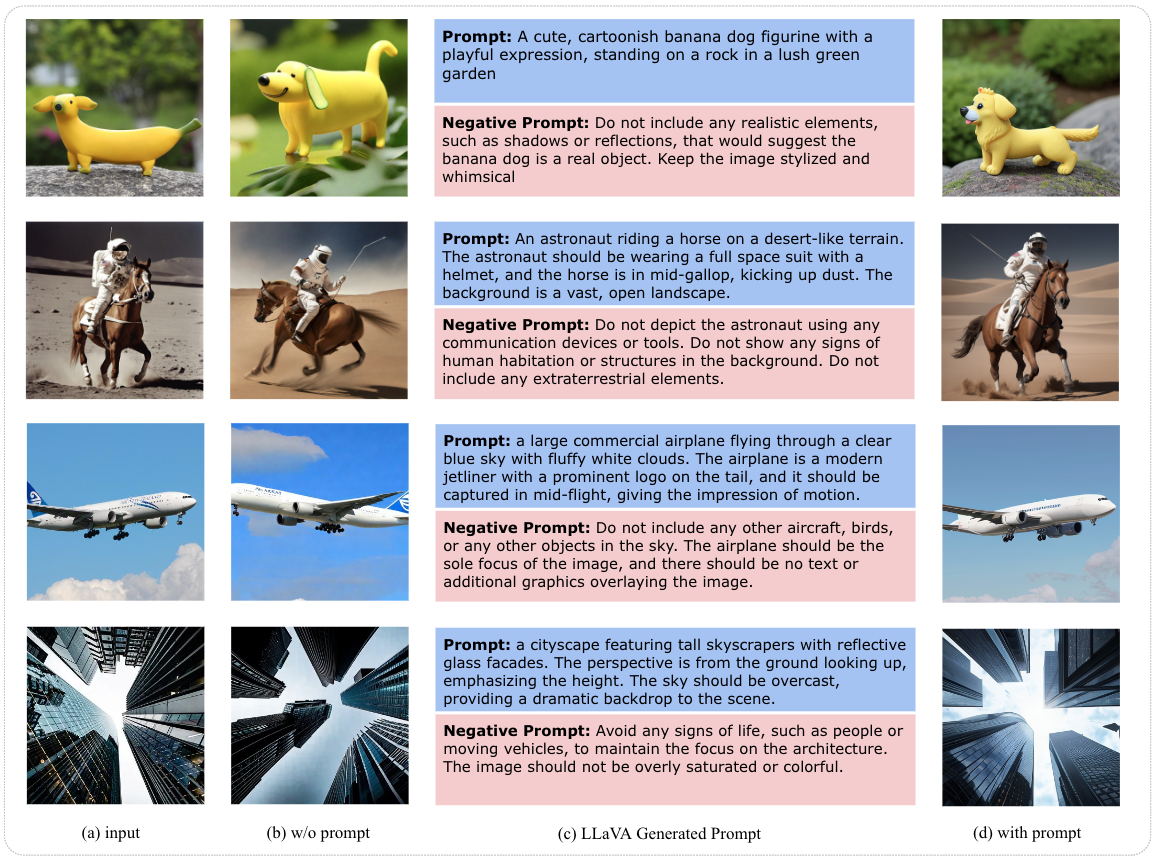}
    \caption{Framework of LLaVA-prompts-based image-to-image generation}
    \label{fig:results}
\end{figure*}

To comprehensively demonstrate the effectiveness of our proposed approach in generating visually similar images, we conducted a supplementary evaluation. This evaluation involved the creation of more input images in different scenarios, encompassing both those generated with the aid of prompts and those produced without. This additional analysis strengthens the generalizability of our proposed approach. Figure 3 demonstrates the images generated with and without LLaVA-generated prompts. Similarly, to better quantify similarity between generated images, we use image similarity metrics, RMSE, PSNR, FSIM, SSIM, UIQ, SRE for comparison. Table 2 showcases that our proposed approach generates more similar images in all these scenarios.

\section{CONCLUSIONS AND FUTURE WORK}
\label{sec:conclusion}

In this work, we propose a novel framework for enhancing image-to-image generation by leveraging the multimodal capabilities of LLaVA. Our approach capitalizes on LLaVA's ability to analyze and comprehend visual content by generating textual descriptions of input image. The LLaVA-generated prompts serve as additional input to the image-to-image generation pipeline. This enriched representation guides the generation process towards outputs that exhibit a better similarity to original image.

Extensive experimentation demonstrates the effectiveness of LLaVA-generated prompts in promoting image similarity. Our results reveal a significant improvement in the visual coherence between the generated image and the input image compared to traditional, prompt-less generation methods.

For the future work, we envision further exploration into fine-tuning LLaVA prompts for increased control over the creative process. By providing more detailed specifications within the prompts, we aim to guide the image-to-image pipeline towards emphasizing specific aspects of the input image while maintaining creative freedom in other areas~\cite{peng2024maxk, peng2023autorep}. This refined approach holds promise for achieving a delicate balance between similarity and creativity.

\renewcommand{\bibfont}{\footnotesize}

\footnotesize{
\bibliographystyle{IEEEtran}
\bibliography{main}

\begin{thebibliography}{10}
\providecommand{\url}[1]{#1}
\csname url@samestyle\endcsname
\providecommand{\newblock}{\relax}
\providecommand{\bibinfo}[2]{#2}
\providecommand{\BIBentrySTDinterwordspacing}{\spaceskip=0pt\relax}
\providecommand{\BIBentryALTinterwordstretchfactor}{4}
\providecommand{\BIBentryALTinterwordspacing}{\spaceskip=\fontdimen2\font plus
\BIBentryALTinterwordstretchfactor\fontdimen3\font minus \fontdimen4\font\relax}
\providecommand{\BIBforeignlanguage}[2]{{%
\expandafter\ifx\csname l@#1\endcsname\relax
\typeout{** WARNING: IEEEtran.bst: No hyphenation pattern has been}%
\typeout{** loaded for the language `#1'. Using the pattern for}%
\typeout{** the default language instead.}%
\else
\language=\csname l@#1\endcsname
\fi
#2}}
\providecommand{\BIBdecl}{\relax}
\BIBdecl

\bibitem{pan2024chain}
Z.~Pan \emph{et~al.}, ``Chain-of-action: Faithful and multimodal question answering through large language models,'' \emph{arXiv:2403.17359}, 2024.

\bibitem{zhang2022can}
R.~Zhang \emph{et~al.}, ``Can language understand depth?'' in \emph{ACM Multimedia}, 2022, pp. 6868--6874.

\bibitem{zeng2024wordepth}
Z.~Zeng \emph{et~al.}, ``Wordepth: Variational language prior for monocular depth estimation,'' \emph{arXiv preprint arXiv:2404.03635}, 2024.

\bibitem{liu2024adaptive100}
H.~Liu \emph{et~al.}, ``Adaptive speed planning for unmanned vehicle based on deep reinforcement learning,'' \emph{arXiv preprint arXiv:2404.17379}, 2024.

\bibitem{shen2024localization}
Y.~Shen \emph{et~al.}, ``Localization through particle filter powered neural network estimated monocular camera poses,'' \emph{arXiv:2404.17685}, 2024.

\bibitem{liu2024using}
M.~Liu, H.~Zhang, J.~Song, and M.~Lu, ``Using generative model for intelligent design of dielectric resonator antennas,'' \emph{Microwave and Optical Technology Letters}, vol.~66, no.~1, p. e34013, 2024.

\bibitem{ding-19-intelligent}
Z.~Ding \emph{et~al.}, ``Using an ancillary neural network to capture weekends and holidays in an adjoint neural network architecture for intelligent building management,'' \emph{arXiv:1902.06778}, 2018.

\bibitem{zhu2024cross}
A.~Zhu \emph{et~al.}, ``Cross-task multi-branch vision transformer for facial expression and mask wearing classification,'' \emph{arXiv:2404.14606}, 2024.

\bibitem{li2024utilizing}
K.~Li \emph{et~al.}, ``Utilizing deep learning to optimize software development processes,'' \emph{arXiv preprint arXiv:2404.13630}, 2024.

\bibitem{ge2023encouraging}
Y.~Ge \emph{et~al.}, ``Encouraging disentangled and convex representation with controllable interpolation regularization,'' in \emph{WACV}, 2023.

\bibitem{zhu2021pseudo}
A.~Zhu, J.~Li, and C.~Lu, ``Pseudo view representation learning for monocular rgb-d human pose and shape estimation,'' \emph{IEEE Signal Processing Letters}, vol.~29, pp. 712--716, 2021.

\bibitem{ge2020zero}
Y.~Ge, S.~Abu-El-Haija, G.~Xin, and L.~Itti, ``Zero-shot synthesis with group-supervised learning,'' \emph{arXiv preprint arXiv:2009.06586}, 2020.

\bibitem{rombach2021highresolution}
R.~Rombach \emph{et~al.}, ``High-resolution image synthesis with latent diffusion models,'' \emph{arXiv preprint arXiv:2112.10752}, 2021.

\bibitem{liu2023llava}
H.~Liu, C.~Li, Q.~Wu, and Y.~J. Lee, ``Visual instruction tuning,'' in \emph{NeurIPS}, 2023.

\bibitem{minaee2024large}
S.~Minaee \emph{et~al.}, ``Large language models: A survey,'' \emph{arXiv preprint arXiv:2402.06196}, 2024.

\bibitem{saharia2022photorealistic}
C.~Saharia \emph{et~al.}, ``Photorealistic text-to-image diffusion models with deep language understanding,'' \emph{arXiv preprint arXiv:2205.11487}, 2022.

\bibitem{pan2023ising}
Z.~Pan \emph{et~al.}, ``Ising-traffic: Using ising machine learning to predict traffic congestion under uncertainty,'' in \emph{AAAI}, 2023.

\bibitem{ma2022learning}
H.~Ma \emph{et~al.}, ``Learning individualized treatment rules with many treatments: A supervised clustering approach using adaptive fusion,'' \emph{Advances in Neural Information Processing Systems}, 2022.

\bibitem{wang2024research}
J.~Wang \emph{et~al.}, ``Research on image recognition technology based on multimodal deep learning,'' \emph{arXiv preprint arXiv:2405.03091}, 2024.

\bibitem{hu2024instructimagen}
H.~Hu \emph{et~al.}, ``Instruct-imagen: Image generation with multi-modal instruction,'' \emph{arXiv preprint arXiv:2401.01952}, 2024.

\bibitem{ramesh2021zeroshot}
A.~Ramesh \emph{et~al.}, ``Zero-shot text-to-image generation,'' \emph{arXiv preprint arXiv:2102.12092}, 2021.

\bibitem{liu2024enhanced}
R.~Liu \emph{et~al.}, ``Enhanced detection classification via clustering svm for various robot collaboration task,'' \emph{arXiv:2405.03026}, 2024.

\bibitem{he2023network}
J.~He, C.~I. Kanatsoulis, and A.~Ribeiro, ``Network alignment with transferable graph autoencoders,'' \emph{arXiv preprint arXiv:2310.03272}, 2023.

\bibitem{jiashu2021performance}
J.~He, ``Performance analysis of facial recognition: A critical review through glass factor,'' in \emph{CDS}.\hskip 1em plus 0.5em minus 0.4em\relax IEEE, 2021, pp. 429--436.

\bibitem{zou2021interpreting}
D.~Zou \emph{et~al.}, ``Interpreting deep learning-based vulnerability detector predictions based on heuristic searching,'' \emph{ACM Transactions on Software Engineering and Methodology}, vol.~30, no.~2, pp. 1--31, 2021.

\bibitem{hang2023efficient}
T.~Hang \emph{et~al.}, ``Efficient diffusion training via min-snr weighting strategy,'' in \emph{ICCV}.\hskip 1em plus 0.5em minus 0.4em\relax IEEE, 2023, pp. 7407--7417.

\bibitem{10402659}
J.~Liu \emph{et~al.}, ``Composable image coding for machine via task-oriented internal adaptor and external prior,'' in \emph{VCIP}, 2023, pp. 1--5.

\bibitem{wang2023diffusiondb}
Z.~J. Wang \emph{et~al.}, ``Diffusiondb: A large-scale prompt gallery dataset for text-to-image generative models,'' in \emph{ACL}, 2023, pp. 893--911.

\bibitem{pan2024cormf}
Z.~Pan \emph{et~al.}, ``Cormf: Criticality-ordered recurrent mean field ising solver,'' \emph{arXiv preprint arXiv:2403.03391}, 2024.

\bibitem{peng2024maxk}
H.~Peng \emph{et~al.}, ``Maxk-gnn: Extremely fast gpu kernel design for accelerating graph neural networks training,'' in \emph{ASPLOS}, 2024.

\bibitem{peng2023autorep}
H.~Peng, S.~Huang \emph{et~al.}, ``Autorep: Automatic relu replacement for fast private network inference,'' in \emph{ICCV}, 2023, pp. 5178--5188.

\end{thebibliography}
}

\end{document}